\documentclass[12pt]{article}

\usepackage{bm}
 \usepackage{natbib}
 \usepackage{amssymb}
 \usepackage{amsbsy}
 \usepackage{amscd}
 \usepackage{amsmath}
 \usepackage{amsthm}
\newtheorem{theorem}{{\bf Theorem}}
\newcommand{\RR}{\mathbb R}

\newcommand{\EE}{\mathbb E}

\title{A Statistical Difference between Single-Layer Learning and Hierarchical Learning in Wide Neural Networks}

\author{Sumio Watanabe\\
sumio.watanabe@riken.jp\\
\\
RIKEN Center for Advanced Intelligence Project
\\
1-4-1 Nihonbashi, Chuo-ku, 103-0027, 
Tokyo, Japan
}

\begin{document}

\maketitle

\begin{abstract}
Hierarchical neural networks are widely used in artificial intelligence, yet their mathematical properties remain incompletely understood. In the infinite-width limit, two different theoretical frameworks have been proposed. One reduces deep learning to kernel regression with a fixed kernel by assuming that the parameters remain close to their initialization, whereas the other allows the parameters to move away from their initialization, requiring the kernel itself to be optimized.

In this paper, we study a three-layer neural network with a finite but large number of hidden units. We show that training the input-to-hidden weights yields a smaller generalization error than keeping them fixed. Furthermore, the latter setting exhibits singularities in the parameter space, whereas the former does not. These findings indicate that singularities play an essential role even in wide neural networks.
\end{abstract}

\section{Introduction}

Inferring an unknown data-generating distribution using a probabilistic model is referred to as machine learning, which has been widely utilized in artificial intelligence and data analysis. In general, the accuracy of such inference improves as the amount of training data increases. To elucidate this behavior, learning theory has been developed \cite{Amari1992, Amari1993,Murata1994},  
and extensive research has been conducted to characterize the learning curve, 
which describes the generalization error as a function of the sample size.
However, learning models with hierarchical structures or hidden variables are referred to as singular models because the correspondence between the parameter space and the model is not one-to-one, and the Fisher information matrix is singular at certain parameter values. Analyzing the learning curves of such models requires methods from algebraic geometry, which is called 
singular learning theory \cite{Watanabe2001, Watanabe2009}. 
Representative examples of singular models include hierarchical neural networks \cite{Watanabe2001ieee, Aoyagi2012, Aoyagi2024, Kurumadani2026}, Gaussian mixture models \cite{Yamazaki2003}, Boltzmann machines \cite{Yamazaki2005}, hidden Markov models \cite{Yamazaki2005ieee}, latent Dirichlet allocation \cite{Hayashi2021}, and factor analysis \cite{Drton2025}.
Building upon these studies, research on achieving AI alignment has recently attracted considerable attention \cite{Wei2022,Hoogland2024}.

Modern artificial intelligence is built upon large-scale hierarchical neural networks, leading to the widespread consideration that the effects of singularities disappear in the large-scale limit. For example, it has been shown that a hierarchical neural network with an infinite number of hidden units in each layer—referred to as an infinite-width neural network—becomes equivalent to kernel regression or Gaussian process regression under an initialization in which the parameters remain within a neighborhood of their initial values \cite{Neal2012, Lee2017, Jacot2018}. Consequently, the learning dynamics are effectively reduced to those of a linear model.
However, practical deep learning does not operate exclusively in such a neighborhood of the initialization. Moreover, it has been demonstrated that this so-called lazy learning regime cannot achieve the performance attained by feature-learning neural networks \cite{Chizat2019, Ghorbani2019, Allen2019}. Therefore, even in infinite-width neural networks, the hierarchical structure remains essential. The effects of singularities do not disappear in practice; instead, they are expected to play an important role in enabling powerful representation learning through phase transitions \cite{Watanabe2001}.

When studying wide neural networks, it is important to recognize that there are many distinct ways of taking the infinite-width limit. In this paper, we compare two learning schemes for a three-layer neural network consisting of an input layer, a hidden layer, and an output layer. In the first scheme, the input-to-hidden weights are randomly initialized and fixed, while only the hidden-to-output weights are optimized during training. In the second scheme, both the input-to-hidden and hidden-to-output weights are optimized.
Although both schemes are universal function approximators in the infinite-width limit, their approximation properties differ substantially, as shown by Barron \cite{Barron1993}. In this paper, we investigate the difference between these two learning schemes from the perspective of statistical inference by analyzing the problem of learning the zero function corrupted by additive random noise. We show that their learning curves remain fundamentally different even in the large width networks. Since the first learning scheme is regular whereas the second is singular, the latter achieves a substantially smaller generalization error despite optimizing a larger number of parameters.

\section{Two Free Energies and Generalization Errors}

In this section, we explain single-layer learning and hierarchical learning.

Let $x \in \RR^M$ and $y \in \RR^N$ be input and output vectors, respectively.
The test sample $(X,Y)$ and the training sample
 $D_n = (X^n, Y^n) = \{(X_i, Y_i); i = 1, 2, \dots, n\}$ 
consist of independent random variables whose probability distribution
 is denoted by $q(y|x)q(x)$, which is referred to as the data-generating distribution or 
 the true 
 distribution. 
The conditional entropy of $q(y|x)$ is given by 
\begin{equation}
S = -\int q(x,y) \log q(y|x) \,dx\,dy. \label{eq:condent}
\end{equation}
The conditional Kullback-Leibler distance of two conditional distributions $q(y|x)$ and $p(y|x)$ is defined by
\begin{align}
K(q(y|x)||p(y|x))& = \EE_{X,Y}\left[ 
\log\frac{q(Y|X)}{p(Y|X)}\right].
\end{align}
Let $p(y|x,a,b)$ be a learning machine which has a parameter $w=(a,b)$.
In this paper we use a prior distribution of $w = (a,b)$, 
\begin{align*}
\varphi(w) &= \varphi_1(a)\varphi_2(b).
\end{align*}
We use two notations, 
\begin{align}
p(Y^n|X^n,a,b) & = \prod_{i=1}^n p(Y_i|X_i,a,b),
\\
q(Y^n|X^n)&=\prod_{i=1}^n q(Y_i|X_i).
\end{align}
The posterior distributions of single-layer learning and hierarchical learning 
are defined by 
\begin{align}
p_1(a|X^n,Y^n,b) &= \varphi_1(a) p(Y^n|X^n,a,b) / p_1(Y^n|X^n,b), \\
p_2(a,b|X^n,Y^n) &= \varphi_1(a)\varphi_2(b) p(Y^n|X^n,a,b) / p_2(Y^n|X^n),
\end{align}
where $p_1(Y^n|X^n,b)$ $p_2(Y^n|X^n)$ are the corresponding marginal likelihoods, 
\begin{align}
p_1(Y^n|X^n,b) &= \int p(Y^n|X^n,a,b) \varphi_1(a) \,da, \\
p_2(Y^n|X^n) &= \iint p(Y^n|X^n,a,b) \varphi_1(a) \varphi_2(b) \,da\,db.
\end{align}
In single-layer learning, $a$ is estimated and $b$ is not estimated, whereas 
in hierarchical learning, both $a$ and $b$ are estimated. 
The predictive distributions of two learning methods
are defined by
\begin{align}
p_1(y|x,X^n,Y^n,b) &= \int p(y|x,a,b) p_1(a|X^n,Y^n,b) \,da, \\
p_2(y|x,X^n,Y^n) &= \iint p(y|x,a,b) p_2(a,b|X^n,Y^n) \,da\,db.
\end{align}
Let 
$\EE_b[\;\;]$ denote the expectation using $\varphi_2(b)$.
The generalization errors are given by
\begin{align}
G_1(n) &= \EE_{X^n,Y^n} \left[\EE_b\left[  K(q(y|x)||p_1(y|x,X^n,Y^n,b))  \right] \right], \\ 
G_2(n) &= \EE_{X^n,Y^n} \left[ K(q(y|x)||p_2(y|x,X^n,Y^n)) \right].
\end{align}
The free energies are also defined as
\begin{align}
F_1(n) &= -\EE_{X^n,Y^n} \left[\EE_b\left[ \log \int p(Y^n|X^n,a,b) \varphi_1(a) \,da \right] \right], \\
F_2(n) &= -\EE_{X^n,Y^n} \left[ \log \iint p(Y^n|X^n,a,b) \varphi_1(a) \varphi_2(b) \,da\,db \right].
\end{align}
By Jensen's inequality, we obtain the following theorem.
\begin{theorem}
For arbitrary $q(y|x)q(x)$, $p(y|x,a,b)$, and $n$, 
\begin{align}
F_1(n) \ge F_2(n). \label{eq:F1F2}
\end{align}
Equivalently, 
\[
\EE_b[K(q(y^n|x^n)||p_1(y^n|x^n,b)]
\geq  K(q(y^n|x^n)||p_2(y^n|x^n)).
\]
\end{theorem}
By definitions,
\begin{align}
p_1(Y^{n+1}|X^{n+1},b) &= p_1(Y_{n+1}|X_{n+1},X^n,Y^n,b) p_1(Y^n|X^n,b),
\\
p_2(Y^{n+1}|X^{n+1}) &= p_2(Y_{n+1}|X_{n+1},X^n,Y^n) p_2(Y^n|X^n),
\end{align}
the following relations hold:
\begin{align}
G_1(n) &= F_1(n+1) - F_1(n) - S, \label{eq:G1F1}\\
G_2(n) &= F_2(n+1) - F_2(n) - S. \label{eq:G2F2}
\end{align}
Therefore, if 
$
F_1(n) - F_2(n)
$
is a non-decreasing function with respect to $n$, then
\begin{align}
G_1(n) \ge G_2(n), \label{eq:G1G2}
\end{align}
since
\[
G_1(n)-G_2(n)=\{F_1(n+1)-F_2(n+1)\}-\{F_1(n)-F_2(n)\}.
\]
It should be emphasized that eq.(\ref{eq:F1F2}), eq.(\ref{eq:G1F1}), 
eq.(\ref{eq:G2F2}), and eq.(\ref{eq:G1G2}) 
hold for an arbitrary data-generating distribution $q(x,y)$ and
 an arbitrary learning machine $p(y|x,a,b)$. 
Note that the numbers of estimated parameters in single-layer learning 
is smaller than that of hierarchical 
learning, however, the free energy and the generalization error of the former are
not smaller than the latter.

Let ${\cal G}_1$ and ${\cal G}_2$ denote the generalization errors of single-layer learning and hierarchical learning, respectively. For $i=1,2$, we decompose the generalization error into the function approximation error ${\cal G}_i^f$, the statistical estimation error associated with the nonredundant part ${\cal G}_i^n$, and the statistical estimation error associated with the redundant part ${\cal G}_i^r$:
\begin{align}
{\cal G}_1 &= {\cal G}_1^f + {\cal G}_1^n + {\cal G}_1^r,\\
{\cal G}_2 &= {\cal G}_2^f + {\cal G}_2^n + {\cal G}_2^r.
\end{align}

The comparison of the function approximation errors of single-layer and hierarchical learning was studied in \cite{Barron1993}. It was shown that, under appropriate smoothness assumptions on the target function, the function approximation error of hierarchical learning is much smaller than that of single-layer learning:
\[
{\cal G}_1^f \gg {\cal G}_2^f.
\]

For the nonredundant part, the statistical estimation error depends on the effective dimension of the parameter space. Since the parameterizations of the single-layer and hierarchical models are different, these errors are difficult to compare directly. Thus, we simply write
\[
{\cal G}_1^n \sim {\cal G}_2^n.
\]
In this paper, we show that the statistical estimation errors associated with the redundant part satisfy
\[
{\cal G}_1^r \gg {\cal G}_2^r,
\]
because of the singularity structure of hierarchical learning.

Since fixing the kernel corresponds to the conventional infinite-width kernel regime, whereas learning the kernel corresponds to feature learning, the above results provide a basis for comparing these two settings. We show that the difference between them becomes more pronounced as the width of the neural network increases. This indicates that the infinite-width limit does not eliminate the distinction between fixing the kernel function and learning it. In other words, fixing the kernel function and learning it are not equivalent, even in the infinite-width limit.

\section{Main Result}

Hereafter we assume that 
a data-generating distribution or a true 
distribution of $y\in\RR^N$ for $x\in\RR^M$ is given by
\begin{eqnarray}
q(y|x)&=& \frac{1}{(2\pi)^{N/2}}\exp\left(-\frac{1}{2}\|y\|^2\right)
\end{eqnarray}
and $q(x)$ is the uniform distribution on $[-1,1]^M$.
In this case, the conditional entropy, eq.(\ref{eq:condent}),
 of the data-generating function is
\[
S=\frac{N}{2}\log (2\pi)+\frac{N}{2}.
\]
We study a learning machine $p(y|x,w)$, which is a neural network with $M$ input units,
$H$ hidden units, and $N$ output units,
\begin{align}
p(y|x,a,b) &= \frac{1}{(2\pi)^{N/2}} \exp\left( -\frac{1}{2} \|y - f(x,a,b)\|^2 \right),
\end{align}
where $(a,b) \in \RR^{HN} \times \RR^{MH}$ and 
\[
f(x,a,b) = \sum_{h=1}^H a_h \tanh(b_h \cdot x).
\]
The prior distribution $\varphi(w)=\varphi_1(a)\varphi_2(b)$ is defined by 
\begin{eqnarray}
\varphi_1(a)&=&  \frac{1}{(2\pi)^{HN/2}}\exp\left(-\frac{1}{2}\|a\|^2\right),
\\
\varphi_2(b)&=& \frac{1}{(2\pi)^{MH/2}}\exp\left(-\frac{1}{2}\|b\|^2\right).
\end{eqnarray}
In sections \ref{section:single} and \ref{section:hierarchical}, 
we show the following theorem. 
\begin{theorem}
Assume that $M,N\ll H\ll n$. Then the generalization errors $G_1(n)$ and $G_2(b)$ of
single-layer learning and hierarchical learning respectively satisfy
\begin{align}
G_1(n)&=\frac{HN}{2n}  + o\left(\frac{1}{n}\right), \label{eq:G1R}
\\
G_2(n)&\leq \alpha(M)
\frac{H^{\frac{M}{M+1}}N^{\frac{1}{M+1}}}{n}
+o\left(\frac{1}{n}\right), \label{eq:G2R}
\end{align}
where
\[
\alpha(M)=\frac{M}{2}
\left(
\frac{2}{(M-1)!}
\right)^{\frac{1}{M+1}}.
\]
\end{theorem}
The coefficient $\alpha(M)$ satisfies $\alpha(1)=\sqrt{2}/2$, $\alpha(2)=2^{1/3}$, and 
when $M\rightarrow\infty$, 
\[
\alpha(M)
=\frac{e}{2}\left(1+\frac{6-\log(2\pi M)}{4M}
+O\left(\frac{(\log M)^2}{M^2}\right)\right).
\]
If $N=1$ and $M$ is large, then this upper bound of $G_2(n)$ 
is not smaller than $G_1(n)$, however, if otherwise, $G_2(n)$ is far 
smaller than $G_1(n)$ in general.
In the special case $M=N=1$, tighter bound 
$G_2(n)\leq \sqrt{HN}/(2n)$ is explained in eq.(\ref{eq:aoyagi2005lambda}). 
Examples of the concrete values for $M=N=100$ and $H=10000$ are given by 
\begin{align}
G_1(n)&\cong \frac{500000}{n},
\\
G_2(n)&\lessapprox \frac{13737}{n}.
\end{align}
Therefore, even for wide neural networks, 
generalization error by learning with a prefixed kernel function 
is larger than learning with an optimized kernel function.

\section{Single-Layer Learning in Neural Networks}\label{section:single}

In this section, we study single-layer learning and show eq.(\ref{eq:G1R}). 
The likelihood function of $a=(a_h^j)\in\RR^{HN}$ is 
\[
p(Y^n|X^n,a,b)=\frac{1}{(2\pi)^{Nn/2}}
\exp\left(
-\frac{1}{2}\sum_{j=1}^N
\left( \sum_{h,k} n I_{hk}(b)a_k^j a_h^j -\sum_h 2B_h^j a_h^j +C^j\right)\right),
\]
where
\begin{eqnarray*}
I_{hk}(b)&=&\frac{1}{n}\sum_{i=1}^n \tanh(b_h\cdot X_i)\tanh(b_k\cdot X_i),
\\
B_h^j&=& \sum_{i=1}^n Y_i^j \tanh(b_h\cdot X_i),
\\
C^j&=&\sum_{i=1}^n \|Y_i^j\|^2.
\end{eqnarray*}
The free energy $F_1(n)$ is obtained by integration of $p(Y^n|X^n,a,b)\varphi_1(a)$ over 
$a$, 
\begin{align}
F_1(n) - nS &=
-\frac{NH}{2}
+\frac{N}{2}
\EE_b\left[
\EE_{X^n}
\!\left[
\log\det J(b)
+\operatorname{tr}(J(b)^{-1})
\right]\right]
\,db,
\end{align}
where
\[
J_{hk}(b)=nI_{h,k}(b)+\delta_{h,k}.
\]
The predictive distribution is concretely given by
\begin{eqnarray*}
p_1(y|x,X^n,Y^n,b) = \frac{1}{(2\pi V^*)^{N/2}} 
\exp\left( -\frac{1}{2V^*} \|y - B^T J(b)^{-1} D(x)\|^2 \right),
\end{eqnarray*}
where $V^*=1+V_1(x)$ and
\begin{eqnarray*}
D_h(x)&=&\tanh(b_h\cdot x),
\\
V_j(x)&=&\mbox{tr}(J(b)^{-j}D(x)D(x)^T). 
\end{eqnarray*}
Then 
the generalization error is given by
\begin{equation}
G_1(n)= \frac{N}{2} \EE^*\left[ \log(1+V_1(X)) - \frac{V_2(X)}{1+V_1(X)} \right],
\end{equation}
where $\EE^*[\;\;]$ is the expectation value over $X$, $X^n$, $Y^n$, and $b$.
By using $V_1(X)\geq 0$ and $x-x^2/2\leq \log (1+x)\leq x$, 
\begin{align}
G_1(n) & \leq \frac{N}{2} \EE^*\left[ V_1(X) \right],
\\
G_1(n) &\geq \frac{N}{2} \EE^*\left[ V_1(X)-V_1(X)^2/2- V_2(X) \right].
\end{align}
Let $\overline{I}(b)$ be the expectation of $I(b)$. 
When $n\rightarrow \infty$, $I(b)\rightarrow \overline{I}(b)$, hence
$J^{-1}\rightarrow 0$, $V_1(X)\rightarrow 0$. Therefore $V_1(X)^2$ and $V_2(X)$ 
go to zero faster than $V_1(X)$. 
\[
G_1(n)=\frac{N}{2} \EE^*\left[ V_1(X) \right] + \mbox{smaller order}.
\]
Let $\{\nu_h(b)\}$ be the set of eigenvalues of $\overline{I}(b)$. 
Then by using the definition, it follows that
\begin{align}
\EE^*[V_1(X)]
&=\EE_b\left[(n\overline{I}(b)(b)+1)^{-1}\overline{I}(b)(b)\right] \nonumber
\\
&=\EE_b\left[\sum_{h=1}^H\frac{\nu_h(b)}{n\nu_h(b)+1}\right] \nonumber
\\
&=\frac{H}{n}-\frac{1}{n}\sum_{h=1}^H
\EE_b\left[\frac{1}{n\nu_h(b)+1}\right].
\end{align}
Then
\[
G_1(n)=\frac{NH}{2n}  + o(\frac{1}{n}),
\]
which shows eq.(\ref{eq:G1R}).

\section{Hierarchical Learning in Neural Networks} \label{section:hierarchical}

In this section, we study hierarchical learning and show eq.(\ref{eq:G2R}). 
In general, the generalization error of this case is given by
\begin{align}
F_2(n)-nS&=\lambda\log n +o(\log n)
\\
G_2(n)&=\frac{\lambda}{n}+o(\frac{1}{n})
\end{align}
where $\lambda$ is the real log canonical threshold \cite{Watanabe2001,Watanabe2009}. 
There exists an algebro-geometric algorithm which enables us to calculate $\lambda$, 
however, it is unknown how $\lambda$ depends on the size of the learning machine. 
\vskip3mm
In the Theorem2 in \cite{Watanabe2001ieee}, the formula of the upper bound 
of the real log canonical threshold 
\[
\lambda \leq \Lambda
\]
was proved, where 
\[
\Lambda\equiv \inf_{0<t<1}
\left\{HMt+\frac{N}{2}\max_{0\leq h}
(h-\Delta(h)t)\right\},
\]
where $\Delta(h)$ is the growing index defined as follows. 
First, we define $\Delta(0)=0$. 
Let binomial coefficient $\binom{n}{k}$ show 
the number of combinations $n$ items taken $k$ at a time. 
For arbitrary positive integer $k$, 
$q \geq 0$ and $0 \leq r < \binom{2q+M}{2q+1}$
are determined so that 
\begin{equation}
k = r + \sum_{p=1}^q \binom{2p+M-2}{2p-1}. \label{eq:qqq1}
\end{equation}
Then $q$ is a monotone non-decreasing function of $k$ and tends to
infinity as $k\rightarrow\infty$. Moreover, if $k\geq M$, then $q\geq 1$, and
if $k<M$, then $q=0$. 
Then the growing index $\Delta(k)$ is defined by
\begin{equation}
\Delta(k) = (4q+2)r + \sum_{p=1}^q (4p-2)\binom{2p+M-2}{2p-1}. \label{eq:qqq2}
\end{equation}
The mathematical reason why $\Lambda$ gives an upper bound of $\lambda$ 
is shown in \cite{Watanabe2001ieee}. It is derived from the inequality 
\[
F_2(n)\leq -\log \int_{\|b_h\|\leq n^{-t}}
\frac{db_1\,db_2\,\cdots db_H}
{\det(n\overline{I}(b)+1)^{N/2}}+\mbox{const.},
\]
where $0<t<1$ is an arbitrary constant and $\overline{I}(b)$ is the expectation of $I(b)$. 
\vskip3mm
\noindent
{\bf Example.} 
In \cite{Watanabe2001ieee}, it was proved that, if $M=1$, then $\Delta(k)=2k^2$,
resulting that $\Lambda=\sqrt{HN}/2$. 
On the other hand, in \cite{Aoyagi2005res}, when $M=N=1$, 
the real log canonical threshold was obtained,
\begin{equation}\label{eq:aoyagi2005lambda}
\lambda=\frac{H+\lfloor\sqrt{H}\rfloor^2+\lfloor\sqrt{H}\rfloor}
{4\lfloor\sqrt{H}\rfloor+2},
\end{equation}
where $\lfloor\sqrt{H}\rfloor$ is the largest integer that is not larger than $\sqrt{H}$. 
These equations show that, in $M=N=1$,  $\Lambda$ gives the tight upper bound 
of the real log canonical threshold $\lambda$. 
\vskip3mm
In this paper, we show eq.(\ref{eq:G2R}) 
by extending this result to
general $M$ and $N$ when $H$ is sufficiently large.

\subsection{Lower bound of the growing index}\label{subsec:2}

In this subsection, we show that the growing index $\delta(k)$ 
satisfies eq.(\ref{eq:Delta(k)geq}). 
By the definition of $q$ and $r$, 
\begin{align}
k & = r +\frac{1}{(M-1)!} \sum_{p=1}^{q} (2p+M-2)\cdots(2p),
\\
\Delta(k) & = (4q+2)r +\frac{1}{(M-1)!} \sum_{p=1}^q (4p-2) (2p+M-2)\cdots(2p). 
\end{align}
If follows that
\begin{align}
k & \leq \frac{1}{(M-1)!} \sum_{p=1}^{q+1} (2p+M-2)\cdots(2p),
\\
\Delta(k) & \geq  \frac{1}{(M-1)!} \sum_{p=1}^q (4p-2) (2p+M-2)\cdots(2p).
\end{align}
By the inequality of arithmetic and geometric means
\begin{align}
k &\leq \frac{2^{M-1}}{(M-1)!} \sum_{p=1}^{q+1} (p+M/4-1/2)^{M-1},
\\
\Delta(k) &\geq  \frac{2\cdot 2^M}{(M-1)!} \sum_{p=1}^q p^{M}.
\end{align}
Bounding the sum of $(p+M/4-1/2)^{M-1}$ and $p^{M}$ from
 above and below using an integral 
\begin{align}
k & \leq \frac{2^{M-1}}{(M-1)!} \frac{(q+M/4+3/2)^M}{M},
\\
\Delta(k) & \geq  \frac{2\cdot 2^M}{(M-1)!} \frac{1}{M+1}q^{M+1}.
\end{align}
It follows that
\[
\frac{\Delta(k)^{1/(M+1)}}{k^{1/M}}
\geq \left(\frac{q}{q+M/4+3/2}\right)
\frac{(2M)^{1/M}((M-1)!)^{1/(M(M+1))}}
{(M+1)^{1/(M+1)}}.
\]
Hence we obtained, 
\begin{equation}\label{eq:Delta(k)geq}
\Delta(k)
\geq C(k) k^{(M+1)/M}
\end{equation}
where 
\begin{align}
C(k) & = C_0(q)C_1(M),
\\
C_0(q)&=
\left(\frac{q}{q+M/4+3/2}\right)^{M+1},
\\
C_1(M)&=
\frac{(2M)^{(M+1)/M}((M-1)!)^{1/M}}
{(M+1)}.
\end{align}
Therefore, if $q\geq M$, 
 $C_0(q)$ is a bounded and monotone increasing function of $q$, 
 which satisfies
\[
1/(1+M/4+3/2)^{M+1}\leq C_0(q)\leq 1.
\]
Thus $C(h)$ is also a bounded and monotone increasing function of $h$. 

\subsection{An upper bound formula of real log canonical threshold}

In this subsection we show eq.(\ref{eq:G2R})
by using eq.(\ref{eq:Delta(k)geq}). By using the definition of $\Lambda$ and 
eq.(\ref{eq:Delta(k)geq}), 
\begin{align}
\Lambda &\leq
 \inf_{0<t<1}
\left\{HMt+\frac{N}{2}\max_{M\leq h}\left\{0,
h-C(h)h^{(M+1)/M}t\right\}\right\}.
\end{align}
Then by setting that $h=s/t^M$, 
\begin{align}
\Lambda &\leq  \inf_{0<t<1}
\left\{HMt+\frac{N}{2t^M}\max_{Mt^M\leq s}\left\{0,
s-C(s/t^M)s^{(M+1)/M}\right\}\right\}.
\end{align}
By choosing a special value 
\[
t^*=
\left[
\frac{N}{2H(M+1)}
\left(
\frac{M}{(M+1)C_1(M)}
\right)^M
\right]^{\frac{1}{M+1}}.
\]
instead of $\inf_{0<t<1}$, it follows that
\begin{align}
\Lambda 
&\leq 
HMt^*+\frac{N}{2(t^*)^M}\max_{M(t^*)^M\leq s}
\left\{0, s-C(s/(t^*)^M)s^{(M+1)/M}\right\}.
\end{align}
When $H\rightarrow \infty$, then $(t^*)^M\rightarrow +0$, hence
$C(s/(t^*)^M)\rightarrow C_1(M)$ uniformly in $M(t^*)^M\leq s$. It follows that, 
for sufficiently large $H$, 
\begin{align}
\Lambda 
&\leq 
HMt^*+\frac{N}{2(t^*)^M}\max_{M(t^*)^M\leq s}\left\{0,
s-C_1(M)s^{(M+1)/M}\right\}. \label{eq:maxs}
\end{align}
Since the function
\[
s-C_1(M)s^{(M+1)/M}
\]
takes the maximum value at
\[
s=\left(\frac{M}{(M+1)C_1(M)}\right)^M, 
\]
the maximum value of this eq.(\ref{eq:maxs}) is exactly calculated as 
\begin{align}
\Lambda & \leq 
(M+1)
(HM)^{\frac{M}{M+1}}
\left[
\frac{N}{2(M+1)}
\left(
\frac{M}{(M+1)C_1(M)}
\right)^M
\right]^{\frac{1}{M+1}} \nonumber
\\
& =
\frac{M}{2}
\left(
\frac{2NH^M}{(M-1)!}
\right)^{\frac{1}{M+1}}.
\end{align}
Thus we obtain eq.(\ref{eq:G2R}).

\section{Discussion}

We discuss the following three points. First, we call
\[
\{\tanh(b_h\cdot x)\;;\;h=1,2,\ldots,H\}
\]
the basis function system. In single-layer learning, this basis function system is randomly fixed before training. In contrast, in hierarchical learning, the basis function system is also optimized as part of the model parameters during training.

First, it has already been known that learning the basis function system according to the unknown data-generating distribution yields a smaller function approximation error than using a fixed basis function system. In this paper, we show that an analogous inequality also holds for the statistical error. From a statistical point of view, as $H$ increases, both methods become capable of approximating arbitrary functions. On the other hand, one might expect that optimizing the basis function system together with the output-layer parameters would increase the generalization error because the number of trainable parameters is larger. However, this does not occur in practice. The same conclusion holds even when the number of layers is increased.

Second, we discuss the effect of the choice of activation function. Equation~(\ref{eq:G1R}) holds whenever the set
\[
\{\sigma(b_h\cdot x)\}
\]
is linearly independent, where $\sigma(x)$ denotes the activation function. Moreover, Equation~(\ref{eq:G2R}) holds if the activation function is an odd function, as is $\tanh(x)$, and admits the Taylor expansion
\[
\sigma(x)=\sum_{p=1}^{\infty} c_px^{2p-1},
\qquad(c_p\neq0).
\]
If the Taylor expansion of the activation function differs from the above form, then the growing index $\Delta(h)$ is also different.

Third, we consider the existence of phase transitions during learning. When the basis function system is fixed, only the parameters connecting the hidden layer to the output layer are updated during training, while the underlying function space remains unchanged. Consequently, even as the number of training samples increases, there is no phase transition in the sense that the support of the posterior distribution changes abruptly.

In contrast, when the basis function system is also learned from data, the log-likelihood function possesses many singularities. Although the posterior distribution takes large values in the neighborhoods of these singularities, the singularities around which the posterior mass is concentrated may change abruptly as the amount of data increases. This means that the effective function space changes with the size of the training data, which is a characteristic feature of hierarchical models. Models that achieve superior performance in both function approximation and statistical inference possess phase transitions in this sense.

\section{Conclusion}

From the viewpoint of statistical inference, we compared single-layer learning and hierarchical learning, and showed that they are not equivalent even for wide neural networks. We also demonstrated that the behavior of learning in the infinite-width limit depends on how the limit is taken. This indicates that, in order to develop a theory that faithfully captures the mathematical properties of hierarchical models, it is necessary to consider limiting procedures that preserve their hierarchical structure.

\bibliography{watabib.bib}

\end{document}